\documentclass{article}

\usepackage{amsmath,amsfonts,bm}

\def\eqref#1{equation~\ref{#1}}
\def\1{\bm{1}}
\newcommand{\train}{\mathcal{D}}

\def\vx{{\bm{x}}}

\def\vz{{\bm{z}}}

\DeclareMathAlphabet{\mathsfit}{\encodingdefault}{\sfdefault}{m}{sl}
\SetMathAlphabet{\mathsfit}{bold}{\encodingdefault}{\sfdefault}{bx}{n}

\def\gL{{\mathcal{L}}}

\DeclareMathOperator*{\argmax}{arg\,max}

\usepackage{arxiv_comp}

\usepackage{lipsum}

\usepackage[utf8]{inputenc} % allow utf-8 input
\usepackage[T1]{fontenc}    % use 8-bit T1 fonts
\usepackage{hyperref}       % hyperlinks
\usepackage{url}            % simple URL typesetting
\usepackage{booktabs}       % professional-quality tables
\usepackage{amsfonts,amsfonts,bm}      % blackboard math symbols
\usepackage{nicefrac}       % compact symbols for 1/2, etc.
\usepackage{microtype}      % microtypography
\usepackage{xcolor}         % colors
\usepackage{xspace} 
\usepackage{array,multirow, graphicx, arydshln}
\usepackage[round, compress]{natbib}

\newcommand{\fewmax}{Few-Max\xspace}
\newcommand{\fastmri}{fastMRI\xspace}
\newcommand{\visda}{VisDA\xspace}
\newcommand{\imagenet}{ImageNet\xspace}
\newcommand{\resnet}{ResNet\xspace}

\title{Few-Max: Few-Shot Domain Adaptation for Unsupervised Contrastive Representation Learning}

\author{%
  Ali Lotfi Rezaabad\\
  The University of Texas at Austin\\
  \texttt{alotfi@utexas.edu} \\
  \And
  Sidharth Kumar\\
  The University of Texas at Austin\\
  \texttt{sidharth@utexas.edu}\\
   \And
  Sriram Vishwanath\\
  The University of Texas at Austin\\
  \texttt{sriram@utexas.edu} \\
 \And
  Jonathan I.\ Tamir\\
  The University of Texas at Austin\\
  \texttt{jtamir@utexas.edu} \\
}

\begin{document}
\maketitle

\begin{abstract}
 Contrastive self-supervised learning methods learn to map data points such as images into non-parametric representation space without requiring labels. While highly successful, current methods require a large amount of data in the training phase. In situations where the target training set is limited in size, generalization is known to be poor. Pretraining on a large source data set and fine-tuning on the target samples is prone to overfitting in the few-shot regime, where only a small number of target samples are available. Motivated by this, we propose a domain adaption method for self-supervised contrastive learning, termed Few-Max, to address the issue of adaptation to a target distribution under few-shot learning. To quantify the representation quality, we evaluate Few-Max on a range of source and target datasets, including ImageNet, VisDA, and fastMRI, on which Few-Max consistently outperforms other approaches. 
\end{abstract}

\section{Introduction}\label{sec: introduction}
A key task in machine learning systems is representation learning \citep{bengio}, as the system performance is often reflected by the quality of the learned representation. In the case of end-to-end supervised training, a (typically) low-dimensional representation is learned either with implicit \citep{dumoulin2016adversarially} or explicit \citep{hasanzadeh2019semi,kingma2013auto,pmlr-v130-lotfi-rezaabad21a}  structure. In some applications, the representation itself is the end-goal, as it offers a concrete way to interpret and visualize high-dimensional data and latent structure \citep{mathieu2019continuous}. These latent representations, while decoupled from specific downstream tasks, can be used more broadly, e.g. for recommender systems \citep{panagiotakis2020unsupervised}, information retrieval \citep{deb2019diversifying}, and as perceptual quality metrics \citep{zhang2018perceptual,wangufloss}. Obtaining sufficient labeled data for training can be challenging in many cases where data collection is expensive, such as in 3D semantic segmentation \citep{csg}, or where it is not possible to obtain labels, such as in magnetic resonance imaging (MRI) reconstruction \citep{wangufloss}.

For this reason, unsupervised representation learning has grown in popularity, owing to its ability to learn rich representations from data without labels and without enforcing explicit structure \citep{wu2018unsupervised,simclr,Moco}. While extremely powerful, these methods are known to be data-hungry, often requiring tens of thousands to millions of examples for training. Pretraining on a data-rich source and fine-tuning on the target domain is a simple approach to adapting the model, and several variants of this basic idea have been proposed, both for self-supervised \citep{csg} and for unsupervised settings \citep{wu2018unsupervised,imix,wang2021unsupervised,wang2020tent,azimi2022self,vu2019advent}. However, performance in the few-shot regime remains poor \citep{shen2021partial}.

Therefore in this work, we consider the scenario of few-shot domain adaptation for unsupervised contrastive learning and aim to improve the quality of the learned representation in a manner that is agnostic to the downstream task. Our approach, which we call \emph{Few-Max}, uses ideas from knowledge distillation \citep{contrastive_distillation} and synthetic-to-real generation \citep{csg} to learn a novel representation using a pretrained network as a regularizer. In addition to including a carefully formed contrastive loss on samples from the target distribution, we guide the learning by enforcing a contrastive loss between the pretrained source-domain network and the target domain network, which are initialized to be the same. To maximally use the limited target domain data, we include an adversarial training objective with simple data augmentation as proposed by \citet{maxup}. We perform experiments on a variety of image-domain data, including few-shot domain adaptation on ImageNet and VisDA, where we evaluate representation quality using quantitative metrics including classification accuracy and diversity of features, and qualitative metrics including image retrieval and loss landscape. We also evaluate \fewmax for adapting a network pretrained on ImageNet to learn representations for complex-valued MRI data, where there are no labels available for training. Such representations have been shown to improve MRI reconstruction accuracy from a limited number of measurements \citep{wangufloss,adamson2021ssfd}. Our results indicate that \fewmax is able to improve the quality of the learned representation with a small number of target domain samples, even when the source domain differs substantially.

Our contributions are the following:
\begin{enumerate}
    \item We propose a novel solution to few-shot domain adaptation for unsupervised contrastive learning without requiring samples from the source domain or labels from the target domain.
    \item We show that our approach leads to a smoother loss landscape, more diversity in the learned representation, and improved generalization to target domains compared to other methods.
    \item We show that our approach can flexibly be used to transfer pretrained representations to completely new domains such as MRI, and that these lead to novel learned feature representations that can be used as perceptual quality metrics in reconstruction.
\end{enumerate}
% 9 Pages of material and as much as supplmentary material. References are extra from 9 pages. Submit the code also. 
% Abstract:- 0.3 page
% Introduction:- 
% Contribution:- 
% Related work:- 
% System Model and Algorithm:- 
% Experiments:- 

\section{Background and Related Work}\label{sec: backgound}
% these are for help to know the references
% is this similar to related works section?
\textbf{Contrastive learning:}
Unsupervised feature learning has substantially improved recently by the use of non-parametric instance discrimination and related contrastive metrics \citep{wu2018unsupervised,simclr,Moco,misra2020self,wang2021unsupervised}. Recent works have shown that these networks serve as better initialization for various downstream tasks including classification, semantic segmentation etc. \citep{chen2020automated}.
Data augmentations that generate semantically meaningful images play a critical role in training contrastive learning methods. The overall objective is to push the representations of augmentations from different examples apart and pull the representations of augmentations from the same examples together to maximize the mutual information between them \citep{tian2020makes,wang2020understanding}. The work in \citet{wu2018unsupervised} used a memory bank to track the features of each example and efficiently calculate the contrastive loss. To support extremely large training sets and augmentations, the work in \citet{Moco} used an exponential moving average to keep track of negative/positive pairs whereas \citet{simclr} generated these pairs through use of extreme data augmentations. The work in \citet{BYOL} tries to directly predict the latent representations of different augmentations thus not requiring explicit negative samples. The authors in 
\citet{ contrastive_distillation} formulate the knowledge distillation problem as as a contrastive learning objective instead the more popular approach of minimizing the KL divergence between the student and teacher networks. 
\citet{adversarial_dp} combined a generative loss with untied weight sharing and discriminative modelling to create a novel framework which generalized prior works related to adversarial adaptation. 
\citet{wang2020tent} proposed an entropy minimization method during test time adaption to optimize the confidence in its outputs. 
\citet{azimi2022self} showed that applying adaptation methods to self-supervised methods during test times could improve generalization. Finally, 
\citet{vu2019advent} showed the application of unsupervised domain adaptation for semantic segmentation using adversarial and entropy losses.

\textbf{Domain adaptation:}
The goal of domain adaptation is to update a model initially trained on a source distribution to adapt to a novel target domain. The adaptation can be fully blind to the target domain or use samples of the target domain to fine tune the source domain pretrained model \citep{csg,gan2016learning,saito2019semi}. The authors in \citet{imix} approached data augmentation from the lens of contrastive learning by training a classifier on unique virtual class labels in each batch. Their classes were formed through linear combinations of different augmentations of the input image, where the mixing coefficient was generated randomly using a beta distribution. Analogous to that, Mixup \citep{mixup}, Cutmix \citep{cutmix}, and Manifold mixup \citep{manifold_mixup} aimed to make the supervised-training neural network robust against memorization and adversarial examples by using different strategies for mixing the input samples. 
Building on the same type of data augmentations, \citet{maxup} proposed to minimize the maximum augmented data loss for each sample rather than its average.
Work in \citet{li2017deeper} proposed to train separate networks for the different source domains and then used the unique parameters from those networks for test time evaluation.  
\citet{chen2020automated} formulated synthetic-to-real image generalization problem as a learning-to-optimize strategy so that the layer wise learning rates are automated. 

\textbf{Few-shot learning:}
In few-shot learning, the objective is to learn a particular task from a limited number of labeled data and then test on newer unlabelled data \citep{cheng2022imposing,chen2019closer,tian2020rethinking}. Generally a model is trained on the source domain and then fine-tuned on a very small dataset from the target domain. Simple transfer learning approaches do not work well owing to the large domain shift and hence the network needs to be properly tuned to optimally work on the target domain. \citet{shen2021partial} proposed to freeze some of the model parameters and fine-tune the rest of the parameters of the model. Different learning rates are used for different layers through evolutionary search. 
\citet{garcia2017few} studied the few shot learning from the perspective of graphical models where both labeled and unlabeled data are present.  
Work by \citet{qi2018low} tried to adapt a pre-trained model by changing the final layer weights as newer categories are added based on embedding layer activations.

\section{Approach}\label{sec: approach}
Given a dataset $\train = \{\vx_i\}_{i=1} ^{N} \subseteq \mathcal{R}^d$, $\vx_i$ represents an instance from the target domain. We assume $f_a$ represents a pretrained ``anchor'' model which is unchanged during the training phase. Accordingly, $f$ is another network that will be optimized for the task of unsupervised contrastive learning. Following the same lines as contrastive learning \citep{simclr}, for each $\vx_i$ we define a set of negative instances $\vx^-_i$. Similarly, we also consider $h_a$ and $h$, two MLP layers, with a pooling operator transformer. With a slight abuse of notation, we assume that $f_{(\cdot)} = f_{(\cdot)} \circ h_{(\cdot)}$. Following the same approach as contrastive learning, one can define the following loss function which potentially can help to transfer the knowledge from a pretrained model $f_a$ to the task model $f$. Therefore, we have
\begin{equation}\label{eq: contrastive}
    \gL_{\text{CL}}(\vx_i) = -\log\frac{\exp(f(\vx_i)^T f_a(\vx_i)/\tau)}{\exp(f(\vx_i)^T f_a(\vx_i)/\tau) + \sum_{\vx^-_{i,k} \in \vx^-_i}\exp(f(\vx_i)^T f_a(\vx^-_{i,k})/\tau)}, 
\end{equation}
where $\tau$ is a temperature hyperparameter ($=0.07$ in this work). In order to maximally leverage the small number of available samples from the target domain in an unsupervised manner, we follow the  {\it CutMix} augmentation strategy \citep{cutmix} to generate an independent set of augmentations $\{\hat{\vx}_{i, m}\}_{m=1}^M$ for each $\vx$. Indeed, for each sample $\vx_i$ in the batch we can randomly choose a sample $j$  within the same batch and mix two inputs as follows:
\begin{equation}\label{eq:cutmix} 
    \text{CutMix}(\vx_i, \vx_j; \lambda) = M_{\lambda} \odot \vx_i + (1 - M_{\lambda}) \odot \vx_j,
\end{equation}
where $\lambda$ is a beta distribution with parameter $\alpha$, $\lambda \sim \text{B}(\alpha, \alpha)$. $M_{\lambda}$ is defined to be binary mask filtering out a region whose relative area is $1 - \lambda$, and $\odot$ is an element-wise multiplication. To simplify the notation, we  define  $\hat{\vx}_i = \text{CutMix}(\vx_i, \vx_j; \lambda)$ (dropping $j$ due to  randomness). Based on this mixing, we can now define the following loss function independently from $f_a$:
\begin{align}\label{eq: con_cutmix}
\begin{split}
       \gL_{\text{task}}(\vx_i) = & -\lambda  \log\frac{\exp(f(\vx_i)^T f(\hat{\vx}_i)/\tau)}{\exp(f(\vx_i)^T f(\hat{\vx_i})/\tau) + \sum_{\vx^-_{i,k} \in \vx^-_i}\exp(f(\hat{\vx}_i)^T f(\vx^-_{i,k})/\tau)} \\ & -(1-\lambda) \log\frac{\exp(f(\vx_j)^T f(\hat{\vx}_i)/\tau)}{\exp(f(\vx_j)^T f(\hat{\vx_i})/\tau) + \sum_{\vx^-_{i,k} \in \vx^-_i}\exp(f(\hat{\vx}_i)^T f(\vx^-_{i,k})/\tau)}.
\end{split}
\end{align}
Intuitively, this contrastive loss can be interpreted as a loss function in which the distance between $f(\hat{\vx}_i)$ and each $f(\vx_i)$, $f(\vx_j)$ has to be minimized and distance between $f(\hat{\vx}_i)$ with negative samples has to be maximized. Importantly, we can have this in-batch blending for each sample $i$ for $M$ times. Therefore, we can get $\{\gL_{\text{task}}(\vx_{i,1}), \gL_{\text{task}}(\vx_{i,2}), \cdots, \gL_{\text{task}}(\vx_{i, M}) \}, ~~ \forall i\in \mathcal{B}$, where $\mathcal{B}$ represents the batch. Finally, we can arrive to our proposed loss function;
\begin{align}\label{eq: distillation contrastive}
\mathcal{L}(\vx_i) = \min_f \Big[ \gL_{\text{CL}}(\vx_i) + \max_{m\in [M]} [\gL_{\text{task}}(\vx_{i, m})]\Big],
\end{align} 
where the minimization occurs on the maximum loss achieved over $M$ in-batch mixture of inputs. Indeed, we can rewrite \eqref{eq: distillation final} as:
\begin{align}\label{eq: distillation final}
\begin{split}
\mathcal{L}_{\text{Few-Max}}(\vx_i) = &\min_f  \frac{1}{N}\sum_{\vx_i}  \gL_{\text{CL}}(\vx_i) +  \gL_{\text{task}}(\vx_{i, m^*}),
\end{split}
\end{align}
where
\begin{align*}
m* = \argmax_m \gL_{\text{task}}(\vx_{i,m}).
\end{align*}
Therefore, one can interpret this loss function for $f$ as a limited adversarial training.

\begin{figure}[h!]
\centering
\includegraphics[width=1\columnwidth, trim={0.2cm 1 1 1},clip]{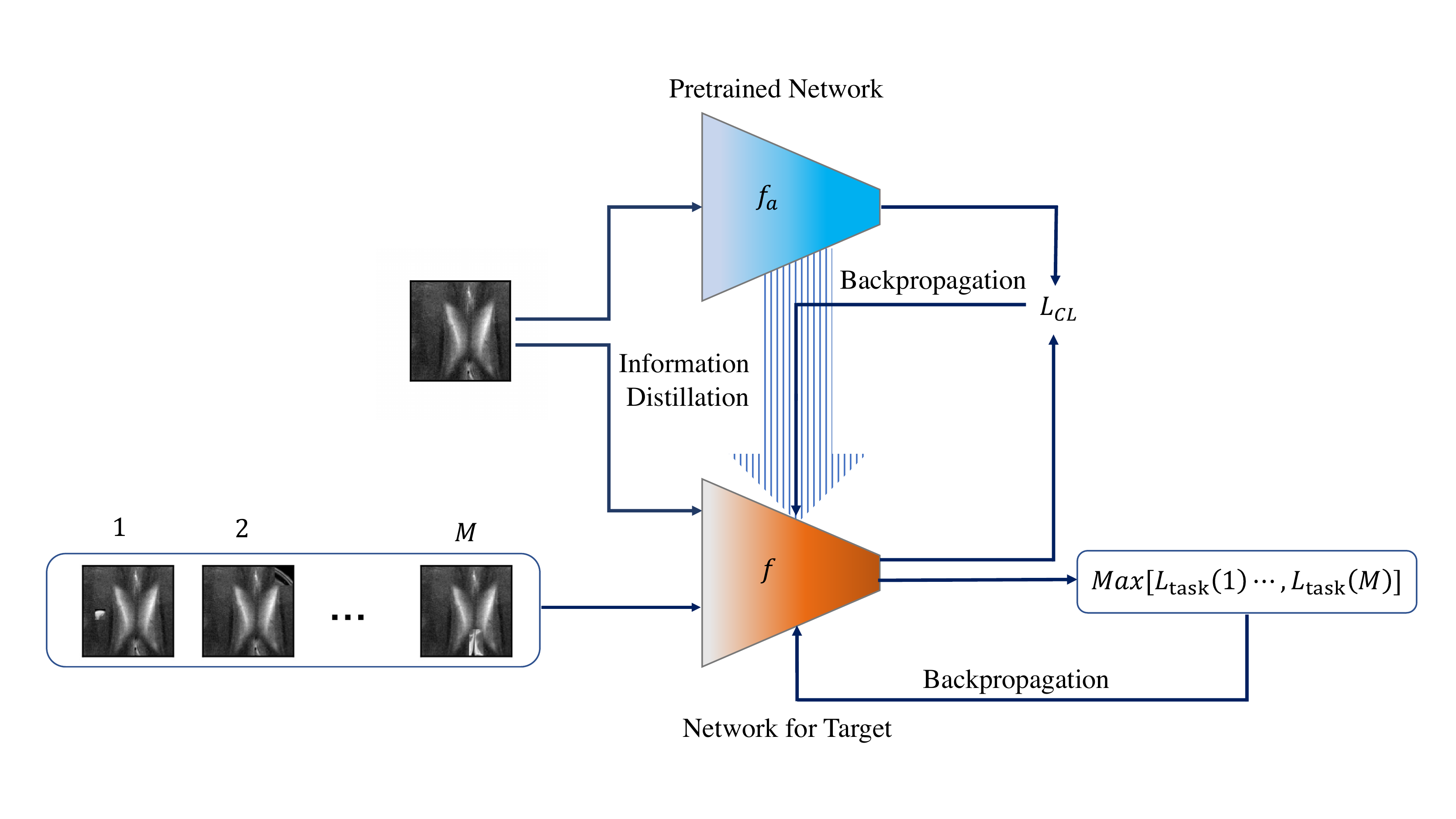}
	\caption{Proposed framework for few-shot domain adaptation for unsupervised contrastive representation learning. $f_a$ represents the pretrained network (shown in blue) which is kept frozen through the entire training procedure. The network $f$ is trained (shown in red) with maximum contrastive loss obtained from $M$ different augmentations of the input in addition to the contrastive loss between the two networks.}
\label{fig: framework}
\end{figure}

Figure \ref{fig: framework} demonstrates the proposed Few-Max architecture. We only assume we have access to target-domain inputs (without labels) and the parameters of the anchor network ($f_a$). Based on these assumptions, this presented work can be interpreted as a test-time domain adaption for unsupervised representation learning. Firstly, a target input and its negative-example inputs pass through both $f_a$ and $f$, and a contrastive loss is computed between them. Secondly, for each epoch, $M$ different augmentations (based on {\it CutMix}) of the same input are obtained by mixing the input with other images within the same batch. Each of the $M$ augmentations independently passes through the $f$, and corresponding contrastive losses are computed. Among these, we select the loss with maximum value from the $M$ losses and the combination of this loss along with the loss of the first step is used to update the network weights.

The proposed  choice of loss function for Few-Max has two distinct benefits. First, it makes the most use of available data by multiple ($M$) blendings of each sample at each epoch. This is crucial as only a few samples are available from the target dataset. It also leads to a smoother loss landscape compared to other methods as we empirically verify in the results. We also provide a theoretical proof for the smoothness of the loss in Appendix A.

\section{Experiments}\label{sec: experiments}
In this section, we demonstrate the effectiveness of \fewmax on several datasets and evaluate the quality of the learned representations via supervised classification on a downstream dataset. We also evaluate the use of \fewmax for adapting a network pretrained on \imagenet to represent complex-valued MRI data from the fastMRI dataset \citep{zbontar2018fastMRI}. We evaluate the quality of the learned representation for knee and brain MRI data via supervised reconstruction loss.

\textbf{Datasets}: (1) \textbf{\imagenet}: \imagenet provides 1.2 million images from 1000 classes for training and 50,000 for validation. For data processing, we used standard data normalization. (2) \textbf{\visda}: the \visda dataset is a large-scale testbed for studying domain adaptation methods, including three splits (domains). Each subset has the same 12 object categories; the training split (source) is collected from synthetic renderings of 3D models from different angles with varying lighting conditions. The validation split includes a real-image domain consisting of images cropped from the Microsoft COCO dataset. (3) \textbf{\fastmri}: we used the \fastmri high-resolution brain and knee data set for our 2D experiments for MRI representation. For each anatomy, data from 10 subjects is included, with 10 central slices from each subject, hence 100 fully sampled 2D MRI images are used. From each slice, 10 patches are extracted, leading to 1000 data points in the few-shot domain adaptation setting. For validation, a total of 5000 patches were used, which were extracted from 50 subjects (10 slices each) for both anatomies. A separate training set consisting of 20000 patches from 200 subjects for each anatomy was created to use for the large-data regime.

\textbf{Methods}: 
{\textbf{Baseline}} represents the conventional unsupervised contrastive learning with random initialization, while {\bf{Finetune}} refers to the same method; however, the network is initialized with a pretrained model instead. {\bf{CSG}} refers to the method by \citet{csg}, which was proposed for labeled synthetic datasets and requires labels for the target domain. {\bf{Un-CSG}} (i.e., unsupervised CSG) refers to the same method, however, with the supervised classification loss replaced with an unsupervised contrastive loss. {\bf{iMix}} refers to the method proposed by \citet{imix}. Finally, {\bf{Few-Mix}} is a special case of our proposed {\bf{Few-Max}}, where $M=1$.

% For each of knee and brain, 100 fully-sampled slices from 10 subjects were used for few-shot domain adaptation, and xx subjects (xx slices) were used for validation.

\textbf{Implementation}: For \imagenet and \visda, we choose \resnet-50 as the backbone. For \fastmri, we used \resnet-18 as the backbone applied to image patches. For training of models, we use SGD optimizer with learning rate of $0.125$, weight decay $0.9\times 10^{-4}$ , and momentum $0.9$. Batch size is set to $64$, and the model is trained for $100$ epochs. The code is available at \href{https://github.com/utcsilab/fewmax}{https://github.com/utcsilab/fewmax}.

\subsection{\imagenet}
We first evaluate our approach in a setting with a large diversity in samples but with fewer examples per class. We emulate this scenario with the \imagenet dataset by randomly selecting 20 classes (labels 151 - 170) as source samples and a different range of 40 classes to form the target dataset (labels 471 - 490). We have 500 samples per class for the source, and 10 samples per class for the target. This study helps us understand the performance of each method in a regime where the target dataset has not only more diverse examples but fewer samples per category. As proposed by \citet{liu2018learning}, one way to measure the generalization is to compute the hyperspherical energy, which evaluates average distances between points in the latent space, with the following expression:
\begin{equation}\label{eq: hyperspherical}
    E_s (\vz) =     \begin{cases}
      \sum_{i\neq j}\lVert \vz_i - \vz_j \rVert^{-s}&  $s > 0$ \\
      \sum_{i\neq j} \log (\lVert \vz_i - \vz_j \rVert ^{-1}) & $s = 0$
    \end{cases}.
\end{equation}

Table \ref{table:imagenet_acc} shows the results of generalization (following \citet{csg}, we report $E_0, E_1, E_2$) and accuracy after adapting a \resnet-50 model to the target dataset with very few samples per category compared to the source. In this experiment, we first train a \resnet-50 on the source dataset for 100 epochs. As mentioned, Baseline and iMix refer to the case where the contrastive model is initialized based on the weights achieved after training on the source using conventional contrastive leaning \citep{simclr} and \citep{imix}, respectively. It is worth highlighting that the key assumption is that we lack any source sample during training on the target, and moreover, neither the source nor the target samples has labels i.e., the training is unsupervised. 

Table \ref{table:imagenet_acc}  shows the evaluation on the basis of the diversity and linear separability of acquired features.  Note that we choose to freeze the backbone network and then a linear classifier is trained on top of the frozen network. In addition, the diversity of features is measured using hyperspherical energy \citep{liu2018learning} (see \eqref{eq: hyperspherical}), and the linear separability is estimated by training a linear classifier on the features. As the results indicate, our proposed Few-Max attained more diverse and accurate features compared to other methods. Finetune and imix refer to the methods by which the pretrained model is optimized on the target samples based on contrastive and imix losses on the target samples, respectively. As the results in Table \ref{table:imagenet_acc} suggest our proposed Few-Mix and Few-Max improves over other methods by achieving more generalizable and diverse features. 
\begin{table}
\centering
\begin{tabular}{l ccc c c}
\toprule
 \multirow{2}{*}{Method} & \multicolumn{3}{c}{Hyperspherical Energy $\downarrow$} &  
\multicolumn{2}{c}{Accuracy $\uparrow$} \\
\cmidrule(lr){2-4} \cmidrule(lr){5-6}
 & $E_0 $& $E_1$& $E_2$ & Top-1 & Top-5\\
 \midrule
\midrule
Finetune    & $0.3267$ & $1.3015$  &  $1.6523$  & $53.05\pm0.1$ & $79.60 \pm0.1$  \\
imix  & $0.3231$ & $1.2830$ & $1.6342$  & $53.55 \pm0.4 $& $79.30 \pm0.3$  \\
Un-CSG  &$0.3049$ & $1.2505$  & $1.5212$  & $55.20 \pm0.5$ & $80.15 \pm0.3$ \\
Few-Mix  &$0.3088$ & $1.2546$  & $1.5357$  & $56.15\pm0.6 $ & $80.60\pm0.4$  \\
Few-Max  &$\textbf{0.2986}$ & $\textbf{1.2457}$ & $\textbf{1.5064}$   & $\textbf{57.20} \pm0.5$& $\textbf{82.05} \pm0.4$\\
\midrule
\midrule
\end{tabular}
\caption{Generalization and diversity of the features extracted by different methods for the \imagenet dataset. Diversity is measured based on hyperspherical energy based on \eqref{eq: hyperspherical} (lower is better) stored in the features. Also, generalization is estimated by assessing a linear classifier on learned features. Further \fewmax gives better Top-1 and Top-5 accuracy, showing that learned features are useful for downstream tasks.}
\label{table:imagenet_acc}
\end{table}

\subsection{\visda}
In this set of experiments, we look at the performance of synthetic data trained model performance on real world samples. Observed by \citet{csg}, the diversity of the learned features plays a key role in model generalization.
To this end, we conduct an experiment where the objective is to transfer the \imagenet pretrained model knowledge to the model trained on synthetic samples of the \visda dataset (without considering labels). It is to be noted that during this experiment we assumed that we have pretrained \resnet model and 1200 unlabeled synthetic samples of the \visda dataset (100 samples per 12 classes).

In Table \ref{table: visda}, we compare different strategies to distill the knowledge from a pretrained \resnet on the \imagenet domain while training on few available unlabeled synthetic samples from the \visda dataset. Like the previous experiment, we quantitatively measure the effectiveness of the models. As a supporting experiment, we evaluated the results based on two different backbone networks. The results suggest our proposed method consistently outperforms the other methods on the basis of diversity and generalization. As a basis for comparison to the state-of-the-art synthetic to real setting, we violate the assumption of lacking labels and provide the results achieved by CSG \citep{csg} in the supervised setting. 

\begin{table}
    \centering
\begin{tabular}{l c ccc c c}
\toprule
 \multirow{2}{*}{Method} &  \multirow{2}{*}{Backbone}& \multicolumn{3}{c}{Hyperspherical Energy $\downarrow$} &  
\multicolumn{2}{c}{Accuracy $\uparrow$} \\
\cmidrule(lr){3-5} \cmidrule(lr){6-7}
 &  & $E_0 $& $E_1$& $E_2$ & Top-1 & Top-5\\
\midrule
Finetune      & \parbox[t]{2mm}{\multirow{5}{*}{\rotatebox[origin=c]{90}{ResNet-50}}} & 0.6708 & 1.4237  & 2.0879  & 40.36  & 80.20  \\
Un-CSG & & 0.5195 & 1.3196  & 1.7914  & 41.20  &  80.44 \\
imix  & & 0.6010 & 1.3756  & 1.9502  & 40.98  & 78.20 \\
Few-Mix  & &  0.5102 &1.3032  & 1.7523 & 41.05  & 81.56 \\
Few-Max  & &$\textbf{0.4883}$ & \textbf{1.2988} & \textbf{1.7347}    & \textbf{42.86} &  \textbf{82.34}\\
\hdashline
CSG  &  & 0.7164& 1.4551 &  2.1765 &  56.60 & 89.14\\
\midrule
\midrule

Finetune      & \parbox[t]{2mm}{\multirow{5}{*}{\rotatebox[origin=c]{90}{ResNet-101}}} & 0.6488 & 1.3835  & 1.9690  & 42.59 & 79.10\\
Un-CSG & & 0.5427 & 1.3774  & 1.8177  & 49.71   &  82.15\\
imix  & &0.5232 & 1.3533  & 1.8149  & 48.78 &  83.41\\
Few-Mix  & & 0.5623 & 1.3812 & 1.8321 & 49.50 & 84.42   \\
Few-Max  & & \textbf{0.5065} & \textbf{1.3098}  & \textbf{1.7604}  & \textbf{51.53} &  \textbf{85.51}\\
\hdashline
CSG  &  & 0.5507& 1.3416 &  1.8559 &  61.13 & 93.1\\
\midrule
\end{tabular}
\caption{Diversity and generalization of learned features across different methods for the VisDA dataset. CSG refers to \citep{csg} where they used labels of synthetic data during adaptation training, as opposed to other methods where we do not leverage labels during the training phase.}
\label{table: visda}
\end{table}

\subsection{\fastmri}
While unsupervised learning has many real-world applications, MRI reconstruction from limited number of measurements is yet another important task that has been shown to benefit from the inclusion of perceptual losses tailored to the data \citep{wangufloss,adamson2021ssfd}. Indeed, the primary idea is to leverage the learned features (achieved via unsupervised contrastive learning) of MRI samples along with deep learning-based reconstruction methods to better maintain perceptual similarity and high-order statistics in the reconstruction. Motivated by these works and, more importantly, considering the cost of collecting fully sampled MRI data, we study the case where there are few available MRI samples for training the perceptual network. To this end, we propose leveraging the abundance of natural image domain data for few-shot learning on MRI data. In this illustration, we utilize the pretrained \resnet-18 model on the \imagenet domain and separately consider two different MRI target domains (i.e., brain and knee images). Following the methodology of \citet{wangufloss}, we extract 10 patches ($90\times 90$ complex-valued) from 100 fully sampled 2D MRI slices (1000 patches in total) and map them to a $128-$dimensional feature vector using the ResNet-18 to distill the knowledge from the \imagenet domain into the MRI domain. We use physically meaningful data augmentations including magnitude scaling and random phase augmentation. We separately train each method for brain and for knee MRI.

\subsubsection{Quantitative Evaluation}
We measure the generalization performance of our proposed method separately for brain and knee MRI data. As the MRI data are intrinsically unlabeled, we need to define a downstream task to provide a quantitative assessment of the quality of the learned features. Inspired by the classification task of \citet{simclr}, we quantitatively evaluate the performance of the features by training a convolutional decoder network to reconstruct the magnitude of the patch from its latent representation. We measure performance in terms of normalized root mean squared error (NRMSE). Our results are reported in Table \ref{table:mri_table}, where baseline refers to the case in which we train a randomly initialized network on the few available samples of the MRI dataset. We also consider the case in which we finetune a pretrained \imagenet on target samples. Our proposed method improves the feature diversity as measured by lower hyperspherical energies in Table \ref{table:mri_table}. It also improves reconstruction quality as measured by NRMSE.
% and structural similarity index (SSIM) \citep{bovikssim}.

\begin{table} 
    \centering
\begin{tabular}{l c ccc c}
\toprule
 \multirow{2}{*}{Method} &  \multirow{2}{*}{Dataset}& \multicolumn{3}{c}{Hyperspherical Energy $\downarrow$} &  
\multirow{2}{*}{NRMSE $\downarrow$} \\
\cmidrule(lr){3-5}
 &  & $E_0 $& $E_1$& $E_2$ & \\
\midrule
Baseline    & \parbox[t]{2mm}{\multirow{4}{*}{\rotatebox[origin=c]{90}{Brain}}} & -0.6369& 0.78451 & 0.6513  & 0.5469   \\
Finetune & &  -0.6245 & 0.7888 & 0.6652  & 0.3799  \\
Few-Mix  & & -0.6748 & 0.7640  & \textbf{0.5834}  & 0.3671   \\
Few-Max  & &\textbf{-0.6711} & \textbf{0.7613} & 0.5908    & \textbf{0.3442}\\
\midrule
\midrule
Baseline    & \parbox[t]{2mm}{\multirow{4}{*}{\rotatebox[origin=c]{90}{Knee}}} & -0.5762 & 0.8199  & 0.7604  & 0.4674   \\
Finetune  & &-0.6068 & 0.8006  & 0.6833  & 0.3338  \\
Few-Mix  & &-0.6021 & 0.7907  & 0.6680  & 0.3213   \\
Few-Max  & &\textbf{-0.6132} & \textbf{0.7862} & \textbf{0.6168}    & \textbf{0.2950}\\
\midrule
\end{tabular}
    \caption{Reconstruction error and diversity of learned features by different methods for the \fastmri dataset. NRMSE refers to the normalized root mean squared error between the magnitude of the ground-truth image patch and the reconstruction using its learned representation as input.}
\label{table:mri_table}
\end{table}

\subsubsection{Qualitative Evaluation} \label{subsec: qualitative evaluation}
\begin{figure}[h!]
\centering
\includegraphics[width=1\columnwidth, trim={0.2cm 29 20 10},clip]{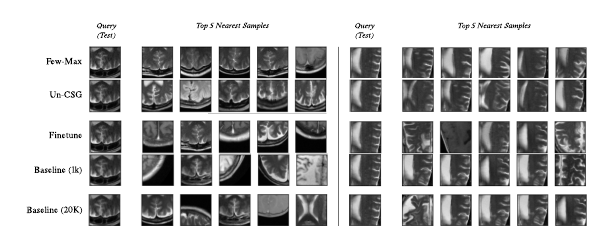}
	\caption{Top-5 retrievals for each test input (query) from the memory bank (consists of 20K inputs) based on distance in feature space. Here, Baseline (1K) and Baseline (20K) refer to training a randomly initialized network with 1K and 20K brain MRI data, respectively. Please see Appendix \ref{App: more results} for more examples.}
\label{fig:image_embedding}
\end{figure}

In this section we follow the methodology of \citet{wu2018unsupervised,wangufloss} and visualize the quality of learned features through latent-space image retrieval. For one evaluation, we build a memory bank of learned features on the large-data training set (e.g. 20K brain patches). Next, we randomly select an input in the test dataset, and we display the top 5 nearest neighbors to this input based on Euclidean distance in feature space. Figure \ref{fig:image_embedding} depicts the results where we can clearly see qualitatively better matches between the test input and the nearest neighbors compared to other methods. For comparison, we also provide the result for the case in which train a randomly initialized model with 1K and 20K image patches, i.e. the large-data regime. For instance, Baseline (1K) and Baseline (20K) are referring to the methods that we train a randomly initialized network with 1K and 20K brain MRI data, respectively. By this experiment, we conclude that our representation is able to capture the salient features in the target distribution in the few-shot regime (more supporting experimental results are provided in Appendix C). Indeed, it reveals the potential of leveraging the abundance of natural images to tailor the representation to other domains such as MRI data, which can be explored more in future work.  

\begin{figure}[h!]
\centering
\includegraphics[width=1\columnwidth, trim={0.2cm 60 20 10},clip]{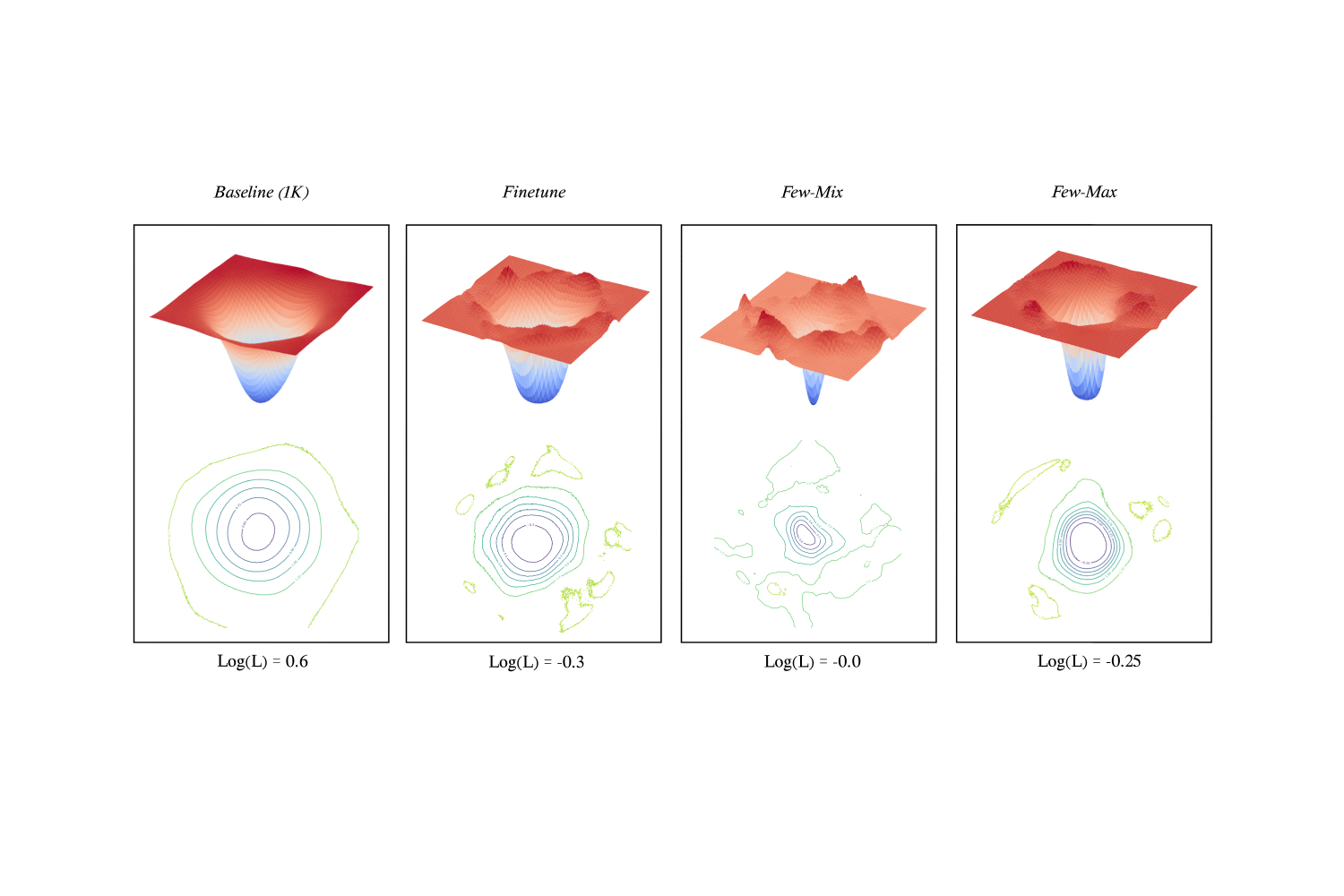}
	\caption{We observe that the loss landscape of Few-Max is smoother than other methods and the actual contrastive loss value have a competitive loss value to the finetune method. It is worth highlighting the loss landscape achieved by \fewmax has less peaks and a smoother landscape around the minimum as compared against \fewmax and finetune method. This is due to the fact the \fewmax loss function minimizes the maximum loss value among different blended images, and it provably imposes a regularizer that minimzes $\lVert \nabla_{\vx} \mathcal{L}_{\text{task}}( \vx)\rVert$.}
\label{fig: loss}
\end{figure}

Using \citet{li2018visualizing}, we also demonstrate the loss landscape assessed on the test dataset for each method on the brain MRI dataset, which supports our claim of the smoothness of the loss. The visualization in Figure \ref{fig: loss} support the claim that  \fewmax has a smoother loss landscape, which helps in explaining its improved generalization performance.
% ali label is below caption dont worry about it
%%%%%%%%%%%%%%%%%%%%%%%%%%%%%%%%%%%%%%%%%%%%%%%%%%%%%%%%%%%%

% It can be clearly observed that the loss landscape of Few-Max technique is much smoother than other methods and moreover the actual loss value is smaller than other methods. This is due to the fact the \fewmax loss function minimizes the maximum loss value among the input batch.
\section{Limitations}\label{sec: limitations}
Our \fewmax approach was only applied to few-shot domain adaptation in the case of balanced data, where each target class is equally represented. It is unclear based on our experiments how the approach will generalize when only unbalanced data are available. This could lead to potential issues related to discrimination, as the quality of the learned features may be biased toward particular over-represented classes in the few-shot regime. In the case of MRI, we only considered learned representations for a fixed patch size and our evaluation criteria used NRMSE. While this metric is known to correlate with radiologist evaluation \citep{muckleyfastmri2020}, a more detailed study is required before the proposed technique can be successfully evaluated for downstream clinical tasks. We also did not evaluate a full MRI reconstruction using the learned features as a perceptual loss, which we leave to future work.

\section{Conclusions}\label{sec: conclusion}
In this paper we presented \fewmax, a general framework for unsupervised few-shot domain adaptation. We use a network pre-trained on a source distribution to initialize a network and guide the training on a target distribution with a small number of unlabeled samples. We use contrastive learning between the source and target networks as a form of regularization, and we use contrastive learning with adversarial domain adaptation to improve generalization to the target distribution. Our experiments show that \fewmax can improve the learned features with a very small amount of target domain data, even when the source and target domains are dramatically different as in the case of \imagenet and \fastmri. The learned features quantitatively and qualitatively capture better representation accuracy as measured by downstream tasks including classification, reconstruction, feature diversity, loss landscape, and image retrieval.
%Bibliography
\bibliographystyle{abbrvnat}
\bibliography{references}

\appendix

\section{Smoothness of Loss}
We investigate how Few-Max loss formulation will introduce a smoother loss landscape by imposing an extra regularizer on $\nabla \mathcal{L}_{\text{task}}$. Expanding upon \eqref{eq: distillation contrastive}, 
\begin{align}
    \begin{split}
    \mathcal{L}_{\text{task, Max}} &= \frac{1}{N} \sum_i \max_{m \in [M]} \big[\mathcal{L}_{\text{task}}( \vx_{i,m})\big]\\
    &=\frac{1}{MN} \sum_i \sum_{m=1}^{M} \mathcal{L}_{\text{task}}( \vx_{i,m}) + \frac{1}{N} \sum_i \max_{m \in [M]} \big[\mathcal{L}_{\text{task}}( \vx_{i,m})\big]\\
    &-\frac{1}{MN} \sum_i \sum_{m=1}^M \mathcal{L}_{\text{task}}( \vx_{i,m})\\
    &=\frac{1}{MN} \sum_i \sum_m \mathcal{L}_{\text{task}}( \vx_{i,m}) + \frac{1}{N}\sum_i\Big[ \max_{m\in [M]} \mathcal{L}_{\text{task}}( \vx_{i,m}) - \frac{1}{M}\sum_m \mathcal{L}_{\text{task}}( \vx_{i,m})\Big].\\
  \end{split}  
\end{align}
Following Lemma 1 in \citet{maxup}, by using Taylor expansion and defining $\mathcal{L}_{\text{task}, \text{avg}} = \frac{1}{MN} \sum_i \sum_{m=1}^M \mathcal{L}_{\text{task}}( \vx_{i,m})$, we can rewrite $\mathcal{L}_{\text{task}, \text{Max}}= \mathcal{L}_{\text{task}}( \vx_{i,m^*})$ as follows,
\begin{align*}
    \begin{split}
        \mathcal{L}_{\text{task}, \text{Max}} = \mathcal{L}_{\text{task}, \text{avg}} + \mathbb{E} \Phi_{\vx} + O(\sigma_2)\\
    \end{split}
\end{align*}
where, 
\begin{align*}
    \begin{split}
   \beta^{-} \lVert \nabla_{\vx} \mathcal{L}_{\text{task}}( \vx)\rVert\leq \Phi_{\vx} \leq\beta^{+} \lVert \nabla_{\vx} \mathcal{L}_{\text{task}}( \vx)\rVert,
    \end{split}
\end{align*}
$\beta^{+}, \beta^{-} \geq 0$ and $\sigma$ depends on the mixing of different inputs samples. Therefore, by the minimization of $\mathcal{L}_{\text{task}, \text{Max}}$ ($\min_f \mathcal{L}_{\text{task}, \text{Max}}$), it imposes a regularizer on $\lVert \nabla_{\vx} \mathcal{L}_{\text{task}}( \vx)\rVert$. It has also been empirically investigated in our experiments, please see Figure \ref{fig: loss}. 

\section{Framework}
Figure \ref{fig: framework} demonstrates the architecture of the proposed Few-Max. We assume only target inputs and the parameters of the anchor network ($f_a$) are accessible throughout this paper. Based on these assumptions, this presented work can be interpreted as a test-time domain adaption for unsupervised representation learning. Firstly, a target input and negative inputs pass through both $f_a$ and $f$, and then a contrastive loss can be computed. Secondly, for each epoch, $M$ different augmentations (based on {\it CutMix}) of the same input also are obtained by mixing the same input with other images within the same batch. Afterward, each $M$ augmentation independently passes the network, and the contrastive losses can be correspondingly computed for them. Finally, we select the loss with maximum value from the $M$ losses and the combination of this loss along with the loss of the first step is backpropagated to update the network weights.

%\section{Implementation Details}
%In our experiments, we use SGD optimizer with learning rate of 0.125. We use ResNet-50 for natural images study case and ResNet-18 for fastMRI one, both followed by the a multilayer perceptron (MLP) project head. We use {\it CutMix} for blending target images within a batch for which $\alpha=1$. The output dimension is 128. The code is available at \href{https://github.com/fewmax/fewmax}{https://github.com/fewmax/fewmax}.

\section{More Results}\label{App: more results}
\begin{figure}[h!]
\centering
\includegraphics[width=1\columnwidth, trim={0.2cm 25 20 10},clip]{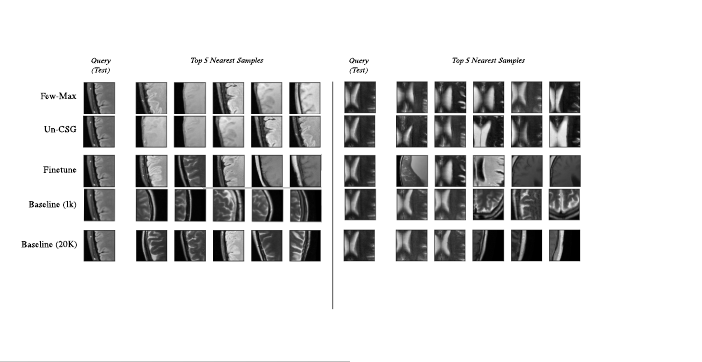}
	\caption{Top retrievals for each test input (query) from the memory bank (training dataset) based on the distance in the feature space. We use \imagenet pretrained network, and the target dataset consists brain MRI samples. Baseline (1K) and Baseline (20k) are the cases that we train a contrastive model with random initialization with which we assume having 1K and 20K samples for the training, respectively. See section \ref{subsec: qualitative evaluation} for more details.}
\label{fig:image_embedding_2}
\end{figure}

Figure \ref{fig:image_embedding_2} illustrates top-5 nearest samples to a test inquiry. Following the same study case as \ref{subsec: qualitative evaluation} , we have the pretrained network on the \imagenet dataset and the target samples are brain MRI data. Clearly, proposed \fewmax enables obtaining a feature embedding network that semantically similar images are closer to each other in that space compared to the others.

\begin{figure}[h!]
\centering
\includegraphics[width=1\columnwidth, trim={0.2cm 25 20 10},clip]{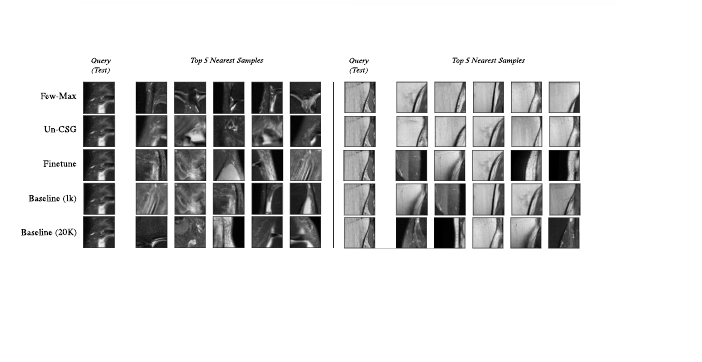}
	\caption{Top retrievals for each test input (query) from the memory bank (training dataset) based on the distance in the feature space. We use \imagenet pretrained network, and the target dataset consists knee MRI samples.}
\label{fig:image_embedding_3}
\end{figure}

\begin{figure}[h!]
\centering
\includegraphics[width=1\columnwidth, trim={0.2cm 30 10 10},clip]{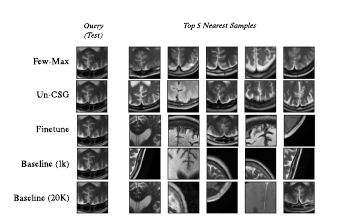}
	\caption{Top retrievals for each test input (query) from the memory bank (training dataset) based on the distance in the feature space. We first train a network on knee MRI dataset (with 20K samples). The target dataset is brain MRI data.}
\label{fig:image_embedding_4}
\end{figure}

Figure \ref{fig:image_embedding_3} represents the results for the scenario,= in which the anchor network ($f_a$) is pretrained on 20K knee MRI data. On the other side, the target samples are brarin MRI data. After training the student network applying different methods, we build a memory bank by 20K brain MRI data.  Figure \ref{fig:image_embedding_4} shows the top-5 nearest for a test inquiry by different methods.  Clearly, Few-Max has an outperforming performance and provides more generalization compared to other methods.

\end{document}